\documentclass[10pt,twocolumn,letterpaper]{article}

\usepackage{cvpr}              

\usepackage{graphicx}
\usepackage{amsmath}
\usepackage{amssymb}
\usepackage{booktabs}
\usepackage{algorithm}
\usepackage{algpseudocode}
\usepackage{xcolor}

\usepackage[pagebackref,breaklinks,colorlinks]{hyperref}

\usepackage[capitalize]{cleveref}
\crefname{section}{Sec.}{Secs.}
\Crefname{section}{Section}{Sections}
\Crefname{table}{Table}{Tables}
\crefname{table}{Tab.}{Tabs.}

\def\cvprPaperID{*****} 
\def\confName{CVPR}
\def\confYear{2022}

\begin{document}

\title{Parking Space Detection in the City of Granada}

\author{Luis Crespo Orti\\
{\tt\small e.luiscrespoorti@go.ugr.es}
\and
Isabel María Moreno Cuadrado\\
{\tt\small e.isa5456@go.ugr.es}
\and
Pablo Olivares Martínez\\
{\tt\small pablolivares1502@gmail.com}
\and
Ximo Sanz Tornero\\
{\tt\small e.ximo@go.ugr.es}
}
\maketitle

\begin{abstract}
   This paper addresses the challenge of parking space detection in urban areas, focusing on the city of Granada. Utilizing aerial imagery, we develop and apply semantic segmentation techniques to accurately identify parked cars, moving cars and roads. A significant aspect of our research is the creation of a proprietary dataset specific to Granada, which is instrumental in training our neural network model. We employ Fully Convolutional Networks, Pyramid Networks and Dilated Convolutions, demonstrating their effectiveness in urban semantic segmentation. Our approach involves comparative analysis and optimization of various models, including Dynamic U-Net, PSPNet and DeepLabV3+, tailored for the segmentation of aerial images. The study includes a thorough experimentation phase, using datasets such as UDD5 and UAVid, alongside our custom Granada dataset. We evaluate our models using metrics like Foreground Accuracy, Dice Coefficient and Jaccard Index. Our results indicate that DeepLabV3+ offers the most promising performance. We conclude with future directions, emphasizing the need for a dedicated neural network for parked car detection and the potential for application in other urban environments. This work contributes to the fields of urban planning and traffic management, providing insights into efficient utilization of parking spaces through advanced image processing techniques.  
\end{abstract}

\section{Introduction}
\label{sec:intro}

Currently, there exists a problem related to the parking spaces in Granada. There is not enough availability leading to conflicts between drivers or spending too much time in order to park. Our project aims to correctly segment satellital images of the city. The objective is to identify the locations of parked cars, moving cars and roads within a given image of Granada. This will be done through different approaches using a neural network model. Given a set of satellite images taken at a certain frequency, it would be possible to extract information about parking areas and their availability over time, providing drivers with tools to park efficiently in the city. For all the code, results and documentation reffer to the project on GitHub\footnote{\url{https://github.com/pab1s/granada-parking-segmentation}}.


\section{Background}

\textit{Fully Convolutional Networks} (FCNs) \cite{long2015fully} represented a significant advancement in the field of deep learning for tasks that require understanding of spatial hierarchies, such as semantic segmentation. The architecture of FCNs typically consist of two parts: the downsampling (encoder) path and the upsampling (decoder) path. In these architectures, the encoder gradually reduces the spatial dimensions of the input image while increasing the depth to extract and learn features at multiple levels of abstraction. This process typically involves a series of convolutional and pooling layers. The decoder, on the other hand, gradually reconstructs the target output from the encoded features, often through a process known as up-sampling or transposed convolution. This architecture is essential for semantic segmentation tasks as it allows for detailed pixel-level predictions while retaining contextual information from the entire image.

\textit{Pyramid networks} were first introduced with the Feature Pyramid Network (FPN) \cite{lin2017feature} and  were designed to solve the problem of capturing objects at various scales. In semantic segmentation and object detection, objects can vary significantly in size, making their detection difficult. Pyramid networks tackle this by implementing a multi-scale approach. They create feature pyramids that maintain information at multiple resolutions, allowing the network to recognize objects and features across different scales effectively. 

\textit{Dilated convolutions} \cite{yu2015multi}, also known as atrous convolutions, provide a solution to the problem of resolution loss in standard convolutional neural networks (CNNs). In semantic segmentation, maintaining high-resolution feature maps is crucial for accurate pixel-level classification. Dilated convolutions enable networks to expand the receptive field of filters without losing resolution or coverage. By adjusting the dilation rate, these networks can aggregate multi-scale contextual information without compromising the resolution. 

\section{Related Works}

It is an established fact that deep learning has significantly advanced the field of computer vision in recent decades. Models such as VGG \cite{simonyan2014very} and ResNet \cite{he2016deep} have demonstrated the capacity to outperform any previous models and have shown exceptional proficiency in extracting features from images. Consequently, these models are often employed as backbones for various semantic segmentation models, including SegNet \cite{badrinarayanan2017segnet}, Mask-RCNN \cite{he2017mask}, PSPNet \cite{zhao2017pyramid}, and DeepLab \cite{chen2017deeplab, chen2017rethinking, chen2018encoder}. However, a significant challenge in deploying these models is their requirement for large amounts of high-quality data \cite{grill2020bootstrap, cordts2016cityscapes}. To address this issue, techniques such as generating synthetic data through generative adversarial networks \cite{ren2020unsupervised} or physics-based models, transfer learning \cite{ahmed2021progressive} or data augmentation \cite{ghaffar2019data} have been proposed as effective strategies for achieving high-quality results with smaller datasets, as suggested in \cite{tang2023semantic}.

A model that has demonstrated proficient results with relatively small datasets is U-Net \cite{ronneberger2015u}. This Fully Convolutional Network (FCN) is specifically designed for medical image segmentation and has gained popularity due to its efficient architecture and effectiveness in handling data with fewer samples. U-Net's architecture is a variation of the encoder-decoder structure which copies the information from the encoder's layers to their symmetric layers of the decoder. This approach has been particularly beneficial for tasks where available training data is limited but high accuracy is required.

Following U-Net's success, several improvements and variations have been proposed. Notably, U-Net++ \cite{zhou2018unet++} introduces a nested, dense skip pathway structure to the U-Net architecture, enhancing the feature propagation and reuse, thus improving the segmentation accuracy. Another notable adaptation is the work by Zhang et al. \cite{zhang2018road}, which modifies the U-Net model for specific applications in road extraction from satellite imagery, demonstrating the model's versatility and adaptability to various segmentation tasks.

Our research focuses on a specific challenge within the realm of semantic segmentation in remote sensing. Detailed aerial imagery provides a wealth of information about urban landscapes, capturing features such as roads, buildings, trees, and vehicles. The design of these urban scenes, governed by stable and consistent constraints, has led to the development of various analytical approaches. Early methods relied on image descriptors, as highlighted by Fua \cite{fua1987using} and Fischler et al. \cite{Fischler1981DetectionOR}, to interpret these complex scenes. Subsequent advancements incorporated texture filters for improved feature extraction, a technique effectively employed in the work of Shotton et al. \cite{shotton2006textonboost}. More recently, the field has seen the integration of diverse machine learning techniques, which have significantly enhanced the ability to analyze and interpret remote sensing data. This evolution is exemplified in the works of Verdie et al. \cite{Verdie2013DetectingPO} and Tokarczyk et al. \cite{tokarczyk2014features}, showcasing the progression towards more sophisticated and nuanced methods of scene analysis. The current state-of-the-art in aerial image semantic segmentation is predominantly driven by advancements in Convolutional Neural Networks (CNNs) and Transformers. The efficacy of CNNs in this domain has been demonstrated through various innovative models that have consistently outperformed their predecessors. A notable example is the work by Sharma et al. \cite{SHARMA201719}, which showcases the application of CNNs in achieving highly accurate semantic segmentation results. Marmanis et al. \cite{marmanis2016semantic} showed that FCNs could achive state-of-the-art results in aerial image segmentation. Additionally, Kaiser et al. \cite{kaiser2017learning} further exemplifies the efficacy of FCN-based approaches in handling aerial imagery segmentation from low quality data from online maps. Recently, He et al. \cite{he2023building} have outperformed state-of-the-art Buildformer \cite{Chen2021BuildingEF} with their Uncertainty Aware Network (UANet).

\section{Methods} 
In the execution of our project, we adopted a methodical and incremental approach.

Our initial objective was to identify the most effective model among three selected candidates for the semantic segmentation of aerial imagery.
The preliminary phase involved training these models using a fusion of two datasets: UDD5 \cite{chen2018large} and UAVid \cite{LYU2020108}, comprising images from China. This amalgamation was strategically chosen to enhance the models' understanding of aerial images, aiming for the network to capture the general characteristics of our problem.

In the second stage, we curated a dataset comprising segmented aerial images of Granada. Leveraging the pre-trained weights derived from the final execution of the top-performing model identified in the initial stage, we conducted retraining with our Granada dataset for several epochs, meticulously refining the model weights. This meticulous process yielded promising outcomes.

To achieve the primary goal of segmenting parked cars from the predicted images, we explored two distinct approaches. The first involved implementing a third stage of post-processing of data, employing a comprehensive heuristic computer vision algorithm method to detect parked cars depicted in \textcolor{red}{Algorithm} \ref{algo:parked-car-detection}.
In summary, the network "Unparked Car Model"  processes parked and moving cars as a single class and it is the heuristic algorithm that later discerns between them.

\begin{algorithm}
\caption{Parked Car Detection}
\label{algo:parked-car-detection}
\begin{algorithmic}[1]
\State Define color codes for car, background and road

\For{each image in the dataset}
    \State Create a mask for the car pixels
    \State Find the contours of the car pixels

    \For{each contour found}
        \State Create a mask for the current contour
        \State Draw the contour on the mask

        \State Create a dilation kernel
        \State Dilate the mask with the kernel

        \State Count the number of background and road pixels within the dilated mask
        \If{background pixels count is greater than road pixels count}
            \State Change color of the car pixels within the original mask
        \EndIf
    \EndFor
\EndFor
\end{algorithmic}
\end{algorithm}
The algorithm employs a dilation operation with a specified 15x15 kernel to enhance car contours, facilitating the connection of disjointed regions. This process effectively smoothens and extends the spatial coverage of the contours, contributing to a broader interpretation of the spatial characteristics.

Alternatively, the second approach to detect parked cars focused on treating parked and non-parked cars as two distinct classes. This modification aimed to empower the network to discern, in its output, which cars are parked and which are not, introducing a new class in the process. This model  "Parked Car Model"  autonomously learn this distinction.

\section{Experiments}

\subsection{Datasets.}

In the training and evaluation of our baseline model, we utilized UDD5 and UAVid datasets. Both datasets consist of images sharing similar characteristics and coherence. Furthermore, we have curated the images by retaining only the classes pertinent to our research, namely roads and cars, while categorizing the remaining elements as 'background'. Therefore, we will have three classes: 'background', 'road' and 'car'.

The project's primary goal was to accurately segment images from Granada. Consequently, it was necessary to apply specific transformations to our data to ensure generalization to this new dataset. Challenges arose from the suboptimal quality of the photos, characterized by low resolution and prevalent shadows, which adversely affected our final model's performance. To mitigate this, a shadow transformation was integrated into the dataloader, resulting in improved model validation with the China images. In addition, we have incorporated various default transformations such as flipping, abrupt rotations, zooming and adjustments in lighting.  For further discussion and examples about this topic see \autoref{domainshift}.

In the final approach, the final model was trained for a few epochs starting from the weights of the baseline model, using a custom dataset named "GranadaAerial"\footnote{The dataset can be found at \url{https://drive.google.com/drive/folders/1rEgcZT_jyJ1zQ88i4epYUKs1QTDDiYya?usp=drive_link}.}, which consists of 90 labeled images for semantic segmentation of parked cars, moving cars and roads from the city of Granada, Spain. This dataset includes 10 images designated for validation and another 10 for testing purposes. The images from Granada used for creating the dataset were obtained from \cite{ignspain_pnoa} according to their policies for educational purposes and segmented using CVAT \cite{boris_sekachev_2020_4009388} and Adobe Photoshop by ourselves.

Regarding dataset partitioning, a manual approach was employed instead of a random split. Given the limited number of segmented images available, we deemed it fair to perform the partitioning manually, thereby ensuring a diverse selection of images in all sets (training, validation and test) as in \cite{Cordts_2016_CVPR}.

Constrained by a limited dataset, we employed data augmentation techniques over the train set, predominantly leveraging flipping, abrupt rotations and lighting adjustments.
Additionally, we introduced an aditional technique enabling image zooming up to 80\% to ensure model consistency with scale among training and validation samples, addressing slight variations in the scales at which the images were captured in the dataset.

\subsection{Baseline Models.} In our study, we selected three baseline models—Dynamic U-Net, PSPNet and DeepLabV3+—each featuring a ResNet101 backbone. These models were chosen based on their architecture characteristics and proven performance, particularly on the Cityscapes dataset, which is closely related to our task of urban scene understanding through aerial imagery.

Dynamic U-Net is an adaptation of the U-Net architecture, known for its effectiveness in aerial image segmentation \cite{aerialUNet}. The 'dynamic' aspect of this model lies in its ability to adjust the architecture's encoder, making it highly adaptable substituting the traditional U-Net encoder by another backbone which can be more appropiate for the actual computer vision problem. This feature is particularly advantageous for our purpose, as it provides to such a powerful model as it is U-Net the capability to do transfer learning, a key aspect considering our computational resources.

PSPNet (Pyramid Scene Parsing Network) has demonstrated exceptional performance in scene parsing tasks, particularly evidenced by its results on the Cityscapes dataset. The model incorporates a pyramid pooling module that works at different scales, enabling it to capture global contextual information effectively. This ability is crucial for semantic segmentation in aerial imagery, where understanding the context is key to accurately classifying various urban elements.

DeepLabV3+, an advanced iteration in the DeepLab series, is renowned for its performance in semantic segmentation tasks, again proven on the Cityscapes dataset. This model introduces an improved atrous convolution strategy and includes an atrous spatial pyramid pooling (ASPP) module, which efficiently captures multi-scale contextual information.

The use of a ResNet101 backbone in the models provides deep feature extraction capabilities across multiple scales, enhancing the models' ability to discern intricate details essential for accurate semantic segmentation in complex urban landscapes.

\subsection{Training Methodology.} The framework chosen to work with the models was Fastai \cite{howard2018fastai}. For uniform training conditions, we employed focal loss \cite{8417976}. In scenarios where certain classes are underrepresented, standard cross-entropy loss may lead to the model being dominated by the majority class, resulting in suboptimal performance for the minority class. Focal loss mitigates this issue by down-weighting well-classified examples, allowing the model to focus more on difficult-to-classify instances. In our first stages we trained PSPNet with cross-entropy obtaining worse results. We employ the Adam optimizer \cite{Kingma2014AdamAM} for all models (Dynamic U-Net, PSPNet, DeepLabV3+), including their ResNet101 backbones pretrained with ImageNet \cite{deng2009imagenet}. The chosen hyperparameters for Adam were $\beta_1 = 0.9$ and $\beta_2 = 0.99$. The training duration was 50 epochs as there was no significant improvement observed across all three models and the metric values stabilized around this point, guided by the one cycle policy for learning rate scheduling \cite{Smith2018SuperconvergenceVF}.

Adopting the one cycle policy as training strategy was based on substantial evidence from numerous studies attesting to its efficacy as widely seen in the literature. Despite its higher resource demands compared to fine-tuning or transfer learning, our hardware capabilities were sufficiently robust to support this approach. In alignment with this strategy, we utilized the learning rate finder method provided by Fastai to ascertain an optimal learning rate, thereby augmenting the overall efficacy of the training process

Each model was assigned a distinct image size, meticulously chosen to maximize the utilization of our available GPU VRAM memory and ensure an equitable comparison. Specifically, we endeavored to maintain a 3:2 image resize aspect ratio whenever feasible, in alignment with the aspect ratios of UDD5 and UAVid images (1.667 and 1.7, respectively). This approach was influenced by precedents set in other studies \cite{xu2018automatic}. A batch size of 32 images was chosen, aligning with findings from various studies that suggest enhanced performance with larger batch sizes in this specific training strategy, as stated by Smith et \cite{smith2018disciplined}.

After comprehensive evaluations, we ultimately chose DeepLabV3+ as our foundational model for the subsequent stage. Further experimentation was conducted with this selected model and the final base model was trained over 60 epochs.

As mentioned earlier, we pursue two different approaches, leading to the training of two distinct models.

The first approach involves initializing our model with the weights obtained from the winning model and training it for twelve epochs using the proprietary dataset from Granada, using only three classes (road, car and background), the model does not differentiate between a parked car and a car in motion.
We executed a limited number of training cycles due to our initial use of pre-trained weights on a Chinese dataset, where the model exhibited satisfactory performance. However, it required adaptation to account for the distinct image scale and road characteristics present in Granada, which differ from those in China. And beyond this number of epochs, there is no substantial improvement in the metric values obtained trying to avoid overfitting.

We use the one-cycle policy. This strategy is designed to efficiently guide the training process, exploiting the cyclical learning rate schedule to enhance model generalization and convergence.

On the other hand, our second approach involves fine-tuning \cite{Shen_Liu_Qin_Savvides_Cheng_2021} the base model using the Granada dataset, distinguishing between moving cars and parked cars. This results in a total of four classes. Consequently, an additional layer is introduced into the base model to accommodate the four-class output

In this case, we undergo a phase of freezing, encompassing the first 5 epochs, during which we focus on training only the model's head while keeping the underlying layers fixed. Subsequently, we will transition to the unfreezing phase, extending for the subsequent 10 epochs.

Furthermore, a batch size of 16 images was chosen in both cases due to, on the one hand, the considerations mentioned earlier about larger batch sizes and on the other hand, because we did not have a very large dataset.

\subsection{Metrics.}
The effectiveness of semantic segmentation models in urban scene understanding, particularly from aerial imagery, is critically evaluated using specific metrics. In this study, we utilize three metrics: Foreground accuracy, Dice Coefficient and Jaccard Index. Each of these metrics offers a unique perspective on model performance.

\begin{table}
  \centering
  \resizebox{\columnwidth}{!}{
  \begin{tabular}{ccccc}
    \toprule
    Model & Valid loss & Foreground acc. & Dice coeff. & Jaccard coeff. \\
    \midrule
    Dynamic U-Net & 0.0736 & 0.6954 & 0.7466 & 0.6269 \\
    PSPNet & 0.0719 & 0.5964 & 0.7240 & 0.5994 \\
    DeepLabV3+ & \textbf{0.05404} & \textbf{0.7726} & \textbf{0.7955} & \textbf{0.6836} \\
    \bottomrule
  \end{tabular}
  }
  \caption{Performance evaluation of segmentation models at the 50th epoch.}
  \label{tab:metric-comparison}
\end{table}

Foreground Accuracy is essential for emphasizing the model's performance in segmenting non-background classes. It can be obtained as follows 
$$ ACC_{foreground} = \frac{TP + TN}{TP + TN + FP + FN} $$
where $TP$ is the number of true positives, $TN$ the number of true negatives, $FP$ the number of false positives and $FN$ is the number of false negatives where ground truth background is not taken into account. This metric is particularly relevant in aerial urban imagery, where the focus is often on diverse urban elements rather than the background. 

Dice Coefficient (DSC) measures the overlap between the predicted segmentation and the ground truth across multiple classes. It computes the Dice Coefficient for each class and averages these values, given as
$$ DSC = \frac{2 |X \cap Y|}{|X|+|Y|} = \frac{2TP}{2TP + FP + FN} $$
where $X$ is the number of segmented pixels, $Y$ is the number of pixels belonging to ground truth. $TP$, $FP$ and $FP$ are the same as for foreground accuracy. This metric is particularly useful for datasets with class imbalances, as it provides equal weight to each class. 

Jaccard Index (JI) is an adaptation of the Intersection over Union (IoU) metric for multiclass scenarios. It calculates the IoU for each class and then averages these scores. It can be calculated from 
$$ JI = \frac{|X \cap Y|}{|X \cup Y|} = \frac{DSC}{2 - DSC} $$
where $X$ and $Y$ are the same as for the DSC. This metric is crucial as it is good handling class imbalance, well-suited for tasks where precise boundary detection and little sensitive to background.


\subsection{Results}

\begin{figure}
  \centering
  \includegraphics[width=0.47\textwidth]{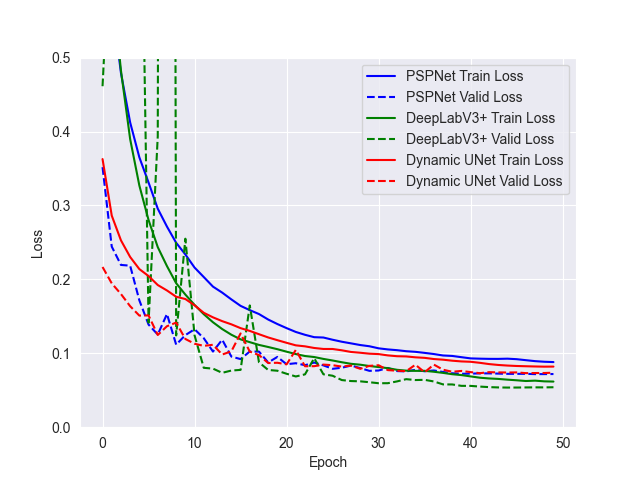}
  \caption{Training and validation loss comparison across epochs for Dynamic U-Net, PSPNet and DeepLabV3+ models. The graph illustrates the trend of validation loss over 50 epochs, highlighting the stability of Dynamic U-Net and PSPNet and the occasional spikes in validation loss for DeepLabV3+, which recovers and maintains a leading performance in subsequent epochs.}
  \label{fig:valid_comp}
\end{figure}

\begin{figure}
  \centering
  \includegraphics[width=0.47\textwidth]{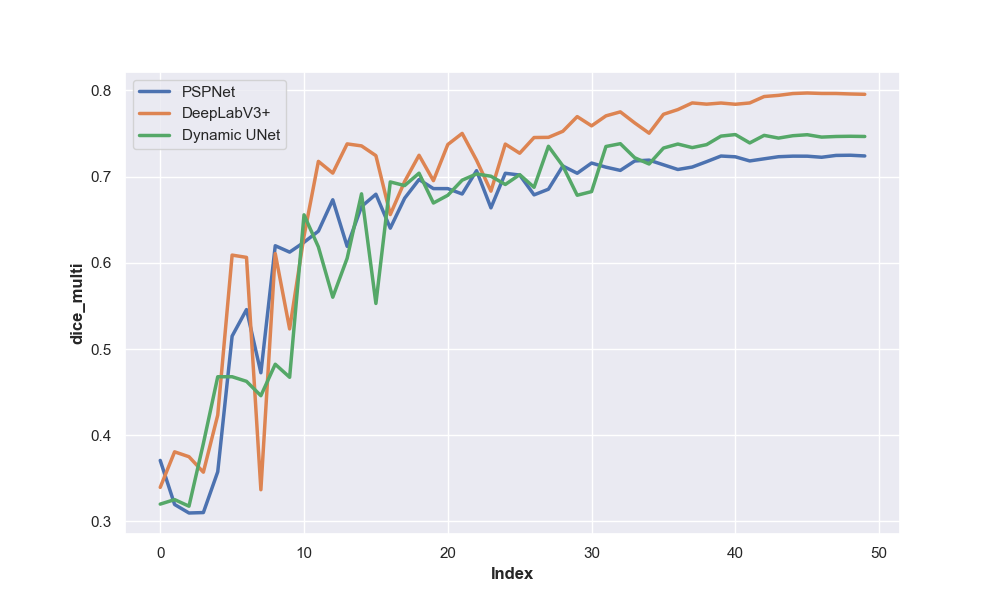}
  \caption{Comparison of Dice Coefficients for Image Segmentation during the first stage of the training.}
  \label{fig:dice_comp}
\end{figure}

After training the three models, we observed a similar pattern of training and validation loss across them, as demonstrated in \autoref{fig:valid_comp}. We extended the training duration despite the slow reduction in validation loss due to consistent improvements across performance metrics, which is particularly notable in the case of DeepLabV3+ as seen in \autoref{fig:dice_comp}. As shown in \autoref{tab:metric-comparison}, DeepLabV3+ achieved the best results among the models, with the lowest validation loss and the highest accuracy and similarity coefficients.

In our study, the validation curve generally stays below the training curve in the graphs. This observation prompts an investigation into the potential influence of the aggressive transformations applied during the data augmentation process on the model's performance.

An inspection of the validation loss graph for DeepLabV3+ reveals intermittent spikes, which we attribute to the model's interaction with more complex or diverse data samples within the batch, reflecting a temporary destabilization in the optimization process, probably due to the aggressive data augmentation accomplished during training. Another hypothesis we considered, as the records of the validation loss\footnote{The records of the metrics can be found in the results folder of the project.} show an isolated abrupt increment of this value in epochs 3 and 8, suggesting a possible overflow of the loss function during those epochs. Nevertheless, despite the feasibility of this hypothesis, we discarded this last option as the training loss behaved as expected. These aberrations underscore the necessity for a delicate balance in learning rate and suggest potential areas for refinement in hyperparameter optimization, despite using the proposed method by Fastai to find an appropiate learning rate for one cycle policy. The resilience of DeepLabV3+ is evident in its rapid recovery following these perturbations, ultimately leading to superior performance metrics.

We now present the outcomes of the two methodologies employed to tackle the Granada problem. 

\begin{figure}
  \centering
  \includegraphics[width=0.47\textwidth]{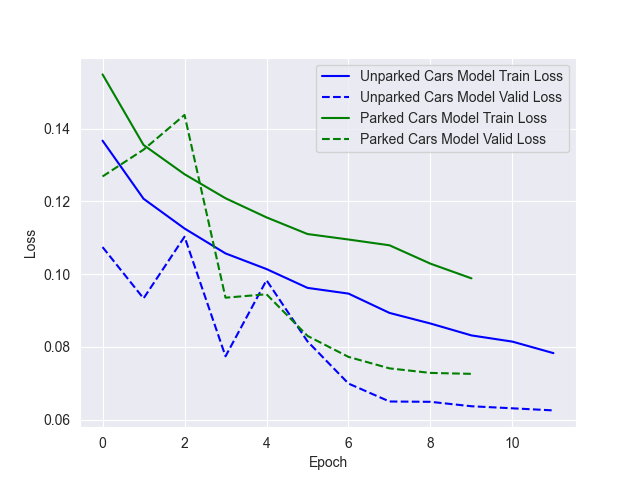}
  \caption{The figure displays the training and validation loss for both the Parked Car Model and the UnParked Car Model.}
  \label{fig:granada_loss}
\end{figure}

In the \autoref{fig:granada_loss} it is shown the train and valid loss of both final models during the second stage. We can assert that the validation loss of the UnParked Car model is reduced to the Parked Car model due to its lower number of classes, resulting in decreased classification complexity.

\begin{figure}
  \centering
  \includegraphics[width=0.47\textwidth]{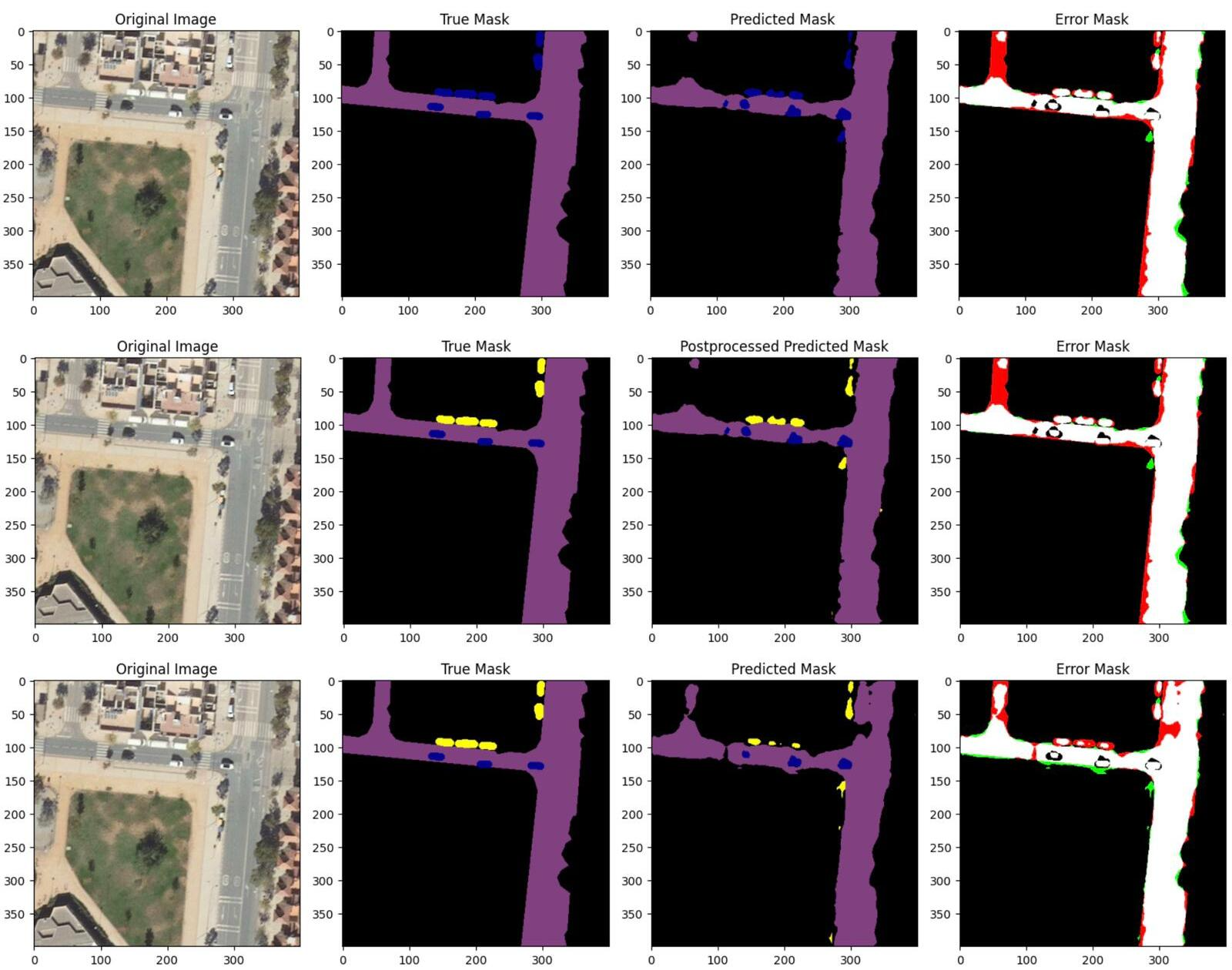}
  \caption{Analysis of Parked Car Detection Methodologies in Aerial Imagery. The analysis showcases three distinct approaches across rows: the baseline detection of cars and roads (Row 1), the implementation of a heuristic algorithm for parked car identification (Row 2) and a model specifically trained to detect parked cars (Row 3). For each method, the columns display the original image, the ground truth segmentation, the algorithm's predicted segmentation and the error mask, respectively. The error mask uses red to signify missed detections (false negatives), green for incorrect detections (false positives) and white for correct detections (true positives).}
  \label{fig:final_results}
\end{figure}

Furthermore, we present two visualizations of the performance of a particular image from the test for both models in \autoref{fig:final_results}.
The first image displays the mask predicted by the first approach, while the following image exhibits the heuristic algorithm for parked car detection applied to that mask.
Similarly the third referees to the second approach. In this instance, as previously mentioned, the model directly predicts areas corresponding to parked cars in the image. 

In both visualizations, white represents true positives, red denotes false negatives and green signifies false positives. Green indicates areas present in the ground truth mask but not predicted, while red indicates areas predicted but not present in the original mask. All evaluations exclude the background.This method was influenced by precedents set in other studies \cite{MESEJO2015167}. 

In this concrete example, each of the two approaches accurately represents the parked cars, as evidenced.

In general, both approaches yield favorable results. This reflects that in specific situations with small datasets, the heuristic approximation can be a viable alternative to a solution entirely based on neural networks.

\section{Conclusions}

In this paper, our primary objective was the detection of parked cars in Granada through semantic segmentation.We pursued this goal by exploring two distinct approaches, both of which have yielded favorable results.

A particular mention should be made regarding the carefully assembled custom dataset of Granada for this study due to the scarcity of datasets specifically designed for locales like Granada within the current research context.

In future work, leveraging daily aerial images of the city of Granada through a suitable tool could enable the use of our model for generating statistics on parking areas. This application extends beyond mere detection, allowing for the analysis of parking behaviors, routines and citizen patterns. Such insights could prove valuable for urban planning and the optimization of parking infrastructure.

To further enhance the capabilities of our model, one avenue for exploration is the expansion of the dataset. By incorporating additional diverse images, including various weather conditions, different times of the day and seasonal variations, we can improve the model's robustness and generalization. This expanded dataset would capture a broader range of parking scenarios, making the model more adept at handling real-world variations.

{\small
\bibliographystyle{ieee_fullname}
\bibliography{egbib}
}

\appendix
\section{Preliminary Model Evaluation Visuals}
\label{domainshift}

\begin{figure}
  \centering
  \includegraphics[width=0.35\textwidth]{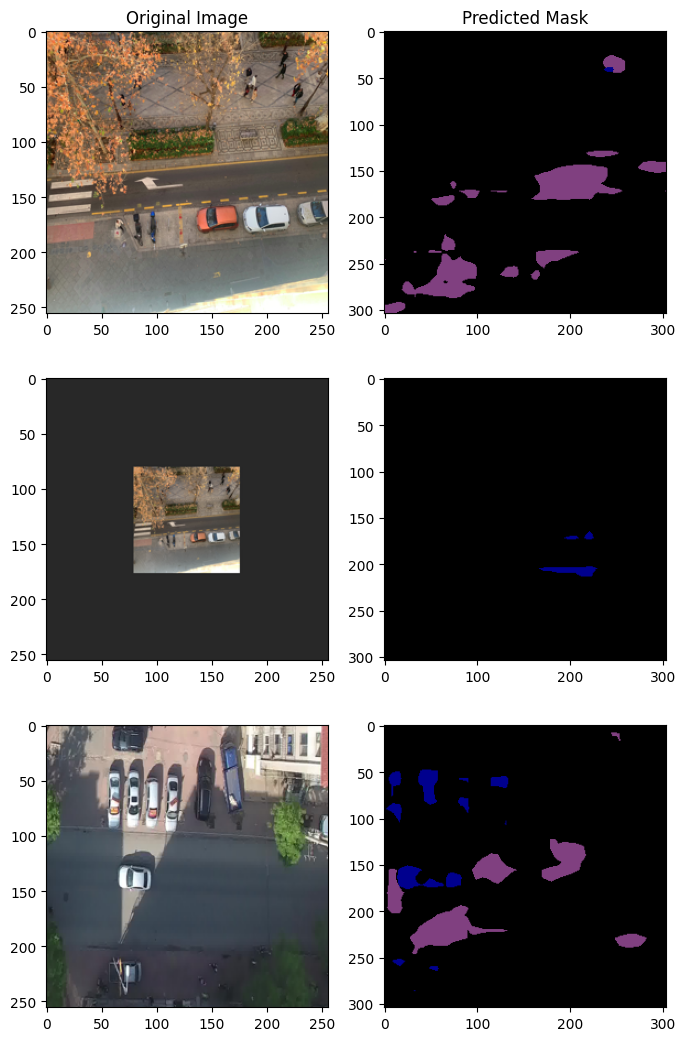}
  \caption{Performance evaluation of our model on images from Granada, exhibiting challenges in identifying cars and roads (left) and the corresponding predicted masks (right). Background is represented with black color, cars in blue color and roads in lilac color. The model's difficulty in detecting roads under shadows and identifying cars and roads at hte same scale suggests a domain shift and limited dataset diversity in the training phase.}
  \label{fig:prel_images}
\end{figure}

In the initial phase of our research, we aimed to evaluate the performance of our model on images from Granada, shortly after the initial training phase. This decision stemmed from an awareness that our training datasets comprised primarily of aerial imagery sourced from countries markedly different from our own. This discrepancy is evident in \autoref{fig:prel_images}, where the model exhibits notable challenges in simultaneously identifying cars and roads within the same frame. Additionally, it struggles to accurately detect roads obscured by shadows. We hypothesize that these difficulties arise from a "domain shift" in the original dataset used for training. Furthermore, the dataset's limited diversity, characterized by wide roads and uniform perspectives, scales and weather conditions, likely contributed to the model's inadequate generalization capabilities.

\end{document}